\RequirePackage{fix-cm}
\documentclass[smallextended]{svjour3}       % onecolumn (second format)
\smartqed  % flush right qed marks, e.g. at end of proof
\usepackage{graphicx}
\usepackage{algorithm2e}
\begin{document}

\title{CSAL: Self-adaptive Labeling based Clustering Integrating Supervised Learning on Unlabeled Data}

%\subtitle{Do you have a subtitle?\\ If so, write it here}

\titlerunning{CSAL}        % if too long for running head

\author{Fangfang Li         \and
        Guandong Xu   \and
        Longbing Cao
}

%\authorrunning{Short form of author list} % if too long for running head

\institute{Fangfang Li \at
              University of Technology Sydney  \\
              \email{Fangfang.Li@student.uts.edu.au}           %  \\
%             \emph{Present address:} of F. Author  %  if needed
           \and
           Guandong Xu \at
             University of Technology Sydney  \\
              \email{Guandong.Xu@uts.edu.au}
           \and
               Longbing Cao \at
               University of Technology Sydney  \\
               \email{Longbing.Cao@uts.edu.au}
}

\date{Received: date / Accepted: date}
% The correct dates will be entered by the editor

\maketitle

\begin{abstract}
%\boldmath
Supervised classification approaches can predict labels for unknown data because of the supervised training process. The success of classification is heavily dependent on the labeled training data. Differently, clustering is effective in revealing the aggregation property of unlabeled data, but the performance of most clustering methods is limited by the absence of labeled data. In real applications, however, it is time-consuming and sometimes impossible to obtain labeled data. The combination of clustering and classification is a promising and active approach which can largely improve the performance. In this paper, we propose an innovative and effective clustering framework based on self-adaptive labeling (CSAL) which integrates clustering and classification on unlabeled data. Clustering is first employed to partition data and a certain proportion of clustered data are selected by our proposed labeling approach for training classifiers. In order to refine the trained classifiers, an iterative process of Expectation-Maximization algorithm is devised into the proposed clustering framework CSAL. Experiments are conducted on publicly data sets to test different combinations of clustering algorithms and classification models as well as various training data labeling methods. The experimental results show that our approach along with the self-adaptive method outperforms other methods.

\end{abstract}

\section{Introduction}
Clustering is an important task in unsupervised learning for dividing objects into different groups. However, the performance of the existing clustering methods is largely limited due to the absence of labels. Differently, classification is advantageous for categorizing the new observations by a supervised classifier learned from a training data set containing labeled observations. The effectiveness of classification, to a great extent, depends on the quality of labeled training data. Actually, lacking labeled data is the core challenging problem in clustering and classification. In real applications, however, it is often very costly and sometimes even impossible to obtain sufficient correctly labeled data for training models. In order to solve this challenge, semi-supervised learning \cite{Demiriz99semi-supervisedclustering} is proposed, which partitions unlabeled data based on the partially available labeled training data, then the original labeled and self-labeled data form a whole training data set for supervised learning. The typical application scenario of semi-supervised learning is when the training data contains a small amount of labeled data but with a large amount of unlabeled data. Semi-supervised learning does not completely solve this problem because it still needs the partially labeled data as a start, which leaves an open question for supervised learning.

As an alternative, recently a new kind of approaches integrating unsupervised and supervised learning on unlabeled data has been proposed, which has shown to be effective in improving the learning performance. Through the new integrated methods, the high quality labeled data points are first produced by purely clustering the unlabeled data and the cluster assignments are used to label training data, and finally the obtained labeled data is involved for better classification. For example, Wang proposed a FCM-SLNMN \cite{WangWCW08} combined algorithm by selecting training data from Fuzzy c-means (FCM) \cite{Dunn-FCM} clustering results. Then the selected training data is applied to build a classifier with a supervised normal mixture model. Similarly, Maulik \textit{et al.} proposed a modified differential evolution (DE)-based fuzzy c-medoids (FCMdd) \cite{MaulikBS10}\cite{SahaMBP11} algorithm. FCMdd selects 50\% of the data points nearest to the cluster center (distance) for training a SVM classifier.

\begin{figure}[htp]
\centerline{\includegraphics[scale=0.4]{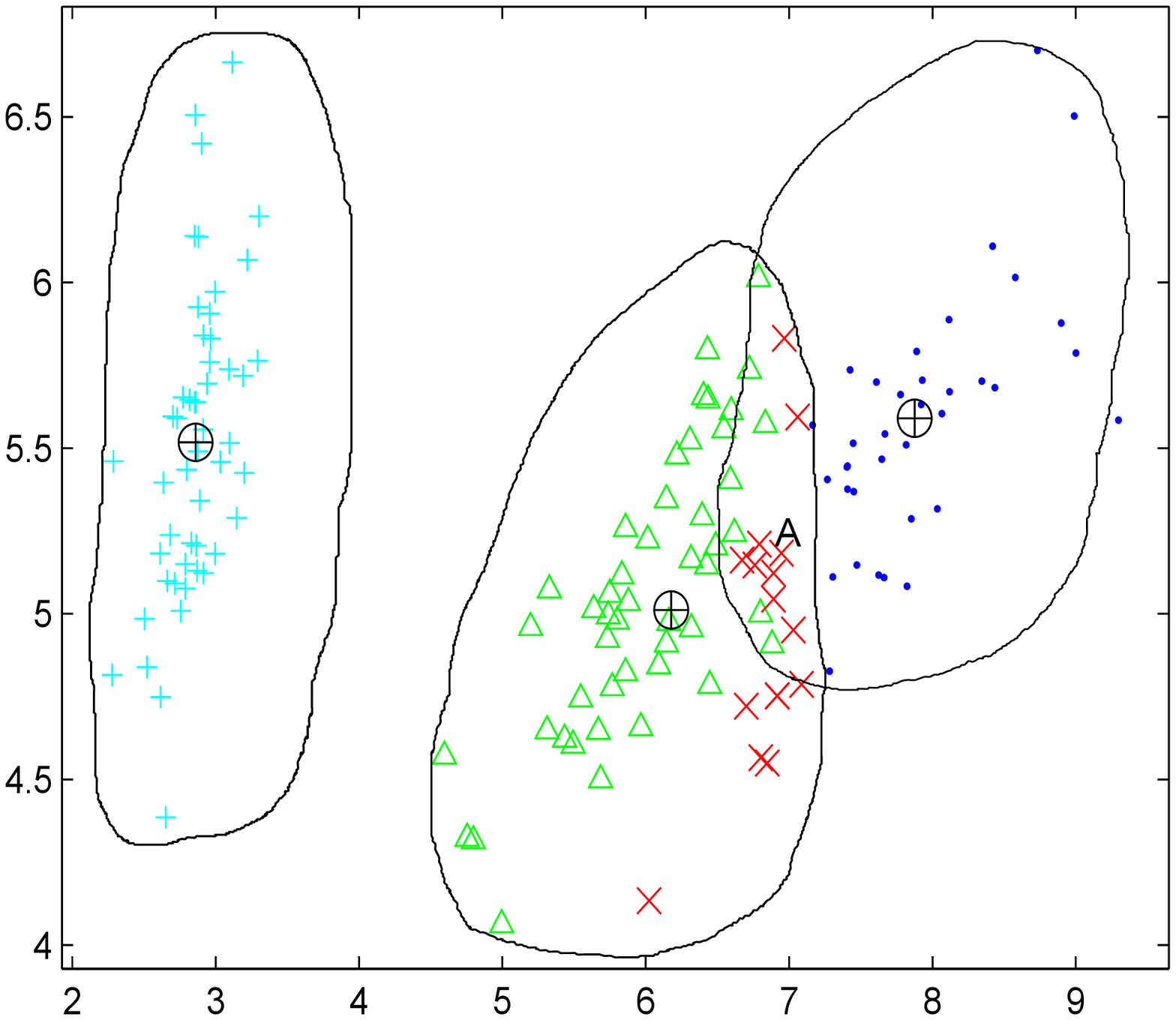}} \caption{The Clustering Results of Iris Data set by K-means. The 4-dimension data is mapped into 2-dimension by PCA. The symbol $\times$ represents the points assigned into wrong clusters, $\oplus$ represents cluster centers.}
\label{fig1_iris}
\end{figure}

Among these methods, the key step is the distance-based labeling process for the unlabeled data (i.e., the correctly assigned data points by clustering algorithms are taken as training data to train a classifier), which essentially impacts the final classification results. Although these methods are more effective than traditional clustering algorithms, they also have some drawbacks. For example, distance-based training data selection method may wrongly label some points on the boundary. Specifically, due to the intrinsic unsupervised limitations of the extant clustering algorithms, the labeling outcome is not perfect: some points are distinctively assigned to one cluster, while others are on the cluster boundary, where the points have the same distance from several cluster centers, are wrongly placed. In order to choose the best quality labeled data, one simple criteria is to select the points which are close to the cluster centers. However, due to the existence of cluster overlapping fact in real applications, there may exist some points which are close to the cluster centers but are located around the cluster boundaries. Fig. \ref{fig1_iris} gives an example of such observations from clustering results on the Iris data set \cite{UCI-Frank+Asuncion} by K-means \cite{Hartigan-Kmeans}\cite{Kanungo0-Kmeans} algorithm. Here although point A is close to the cluster center, it should be excluded from the labeling.  It indicates that simply relying on distance-based methods may select wrong data points into the training data, and eventually mislead the final partition performance.

Distance-based labeling algorithm chooses the data points which are close to cluster center point. The algorithm works reliably when most points are closely gathered in different groups. However, this approach fails to handle the clusters which possess some intersections, i.e., the cluster boundaries are overlapping largely. In another word, some points close to cluster centers but around the boundaries should not be labeled. In order to solve this problem, entropy-based algorithms \cite{MinimusEntropy} are proposed which estimate the entropies of data points and select the points with lower entropy values, indicating the points very likely belonging to one cluster but weakly associated with other clusters. This approach can easily identify the points around cluster boundaries than the distance-based method. In short, the entropy-based selection approach handles the points around the boundaries well, while the distance-based labeling approach finds the good points close to the clustering center well. However, due to the diversity of cluster characteristics which undoubtedly influence the labeling method performance, only one labeling approach is not able to manipulate the various cluster characteristics. In this paper, we thus propose a self-adaptive labeling approach for selecting training data in line with the cluster characteristics. The underlying idea is that if a cluster is compact and separated well from others, distance-based labeling will be applied in the clusters. Otherwise, if one cluster is intersected and not separated well with other clusters, the entropy-based labeling will be employed. In particular, we adopt Silhouette coefficient \cite{dataClusteringJain10} to measure the compactness and separation of the clusters and determine which method should be applied.

Although the labeling approach can find possible labeled training data, the classification is not always satisfied as a result of the quality of labeled training data, derived from the clustering. For instance, data point $x_i$ which belongs to cluster $C_m$ is wrongly partitioned into cluster $C_n$ ($m \neq n$). This scenario will result in the wrong labels for the training data. Thus the classification performance will be accordingly limited. This problem motivates us to further devise a solution following a labeling-classifying-relabeling-reclassifying mechanism to refine the classifier training.

Therefore, we propose an iterative clustering framework by combining the above two strategies, i.e., self-adaptive labeling and classifier refinement. The iterative process aims at refining the classifiers from labeled training data via an expectation-maximization algorithm. In each iteration, we use the trained classifiers to classify the source data and all the labels of the source data are updated accordingly, then the improved labels of training data are fed back to train the classifiers iteratively until the convergence of classifications. Unlike the semi-supervised learning, the self-adaptive labeling based clustering (CSAL) method does not need any labeled data in advance, instead it only involves clustering for labeling training data and building classifiers.

The contributions of the paper are as follows:
\begin{itemize}
\item{We propose the CSAL method which integrates clustering and classification together for unlabeled data.}
\item {We compare the self-adaptive labeling method with distance-based and entropy-based labeling approaches.}
\item{We conduct substantial experiments to verify our labeling algorithms and integrated framework.}
\end{itemize}

The rest of this paper is organized as follows. Section 2 presents the related work. In Section 3 we first introduce the proposed clustering framework, then detail training data labeling algorithms. Section 4 introduces several clustering algorithms integrating with supervised classification together such as CEM and CSAL. Experiments are then conducted and analyzed in Section 5. The paper is concluded in the last section.

\section{Related Work}
\subsection{Semi-Supervised Learning}
The underlying idea of semi-supervised learning is taking the existing rare labeled data as a guidance to partition unlabeled data for reducing the painful manually labeling process. Blum and Mitchell proposed a co-training approach \cite{BlumM98-co-training} which uses two independent classifiers and labeled examples to classify the unlabeled data and pick up the most confident positive and negative examples to enlarge the training set of the other. Nigam and Ghani \cite{Nigam-co-training} proposed co-EM which combines the co-training and Expectation Maximization (EM) for low errors. Nigam \textit{et al.} \cite{semi-EM2} also introduced an algorithm for learning from labeled and unlabeled documents based on the combination of EM and a Naive Bayes classifier. Some other good outcomes such as graph-based semi-supervised learning \cite{graph-basedSL} \cite{LiuWC12-GBSL}, semi-supervised support vector \cite{Bennett98semi-supervisedsupport} \cite{Li-improvingSemiSVM} and multi-manifold semi-supervised learning \cite{GoldbergZSXN09-multissl} also assumed that a small amount of labeled data is involved. Generally, semi-supervised learning can be typically used when the data has small pieces of labeled data with a large amount of unlabeled data. This indicates that a few data points needed to be labeled first when applying semi-supervised learning approaches on the completely unlabeled data. In real applications, however, the labeling is often time-consuming. So it is beneficial to integrate clustering with classification for directly handling unlabeled data.

\subsection{Integrating Clustering for Classification}

In order to solve the challenging problems, the combination of clustering and classification becomes an active research area. Unlike semi-supervised learning combining labeled data with unlabeled data together, this method integrates clustering and classification for directly partitioning on unlabeled data. For example, Celeux and Govaert \cite{Cel92-CEM} described a classification EM algorithm (CEM) which is a classification version of the EM algorithm, it incorporates a classification step between the E-step and M-step of the EM algorithm by a maximum a posteriori principle. Wang also proposed a FCM-SLNMN clustering algorithm \cite{WangWCW08} by a distance-based training data labeling method from FCM clustering results. But the performance of the extant integrated classification algorithms is limited by distance-based training data labeling approach. Different from FCM-SLNMN, the proposed CSAL relies on a self-adaptive labeling process considering information entropy and distance to select training data, leading to much more reliable labeled data for further classification.

\section{CSAL Framework}

\begin{figure}[htp]
\centering
\includegraphics[scale=0.7]{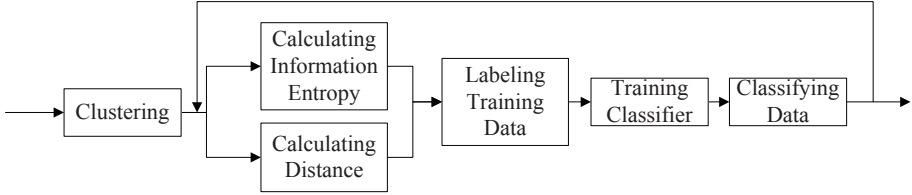}
\caption{The CSAL Framework} \label{fig_framework}
\end{figure}

This section introduces the effective clustering framework (CSAL) which is described in Fig \ref{fig_framework}.

\subsection{Notations}

For better illustration of the CSAL framework, we first list the notations used in this paper in Table \ref{tab-liu2}.
\begin{table*}[htbp]\caption{Notations}
\centering
\begin{tabular}{|l|l|}
\hline
\multicolumn{1}{|c|}{Notations}% time}
& \multicolumn{1}{|c|}{Explanations} \\% cost}\\
%\hline
 \hline
 $x=\{x_{1},x_{2},...,x_{N}\}$  & input data set including $N$  points\\ \hline
 $x_{i}=\{x_{i1},x_{i2},...,x_{id}\}$  & $d$-dimension vector\\ \hline
 $C_{l}$  &  a specific cluster, $l=1,...,K$\\ \hline
 $P(C_{l}|x_{i})$  &  probability of a point $x_{i}$ to be assigned to cluster $C_{l}$, \\ \hline
 & where $l=1,...,K, i=1,...,N$.\\ \hline
 $H(x_{i})$  &  information entropy of a point $x_{i}$, $i=1,...,N$.  \\ \hline
 $s(x_{i})$  &  silhouette coefficient $s(x_{i}$) of point $x_{i}$.  \\ \hline
 $MS(C_{l})$  &  mean silhouette coefficient  \\ \hline
 $E$  &  selected training data  \\ \hline
 ${\omega_{il}}^k$  &  posterior probabilities of $x_{i}$ belonging to $C_{l}$ \\ \hline
 $\theta=\{\alpha_{l},\mu_{l},\Sigma_{l}\}$  &  parameters to train a classifier, where $\alpha$, $\mu$, $\Sigma$ respectively \\
 & represent the mixing probability, means and covariance.\\ \hline
 $A\%$  &  parameter for controlling the percentage of selected data.  \\ \hline
 $s(x_i)$  &  the silhouette coefficient of point $x_i$.  \\ \hline
 $MS(C_l)$  &  the mean silhouette coefficient for a specific cluster $C_l$.  \\ \hline
\end{tabular}
\label{tab-liu2}
\end{table*}

\subsection{The CSAL Framework}

The core components of the CSAL framework include: clustering, training data labeling, training classifier and iteration. Among the components, training data labeling plays a vital role by connecting clustering and supervised classifier together. Since the selected training data may have wrong assigned points, the reselecting and iteration steps aim to refine the selected training data after considering the classification results. The process of CSAL framework is detailed below:

\begin{enumerate}
\item	Clustering: Clustering algorithms are employed to divide the unlabeled data into different groups;
\item	Calculating information entropy and distance: Information entropy and distance are computed for every point in terms of its probability of being associated with each cluster;
\item	Labeling training data: Different strategies (distance-based, entropy-based and self-adaptive methods)  are applied to select training data;
\item	Training a classifier: Classifiers are trained based on the training data set;
\item	Classifying data: The trained classifier is used to classify the source data;
\item	Recalculating information entropy and distance: Information entropy and distance are recalculated regarding the classification outcomes;
\item	Reselecting training data: Training data is refined based on the results of step 6);
\item	Iteration: Repeating the steps 2) - 7); and
\item	Stopping: The algorithm stops until the partition is converged.
\end{enumerate}

\subsection{Training Data Labeling Algorithms}
The goal of clustering is to group $N$ data points into $K$ clusters. To improve the clustering results, the selected training data should consist of a proper quantity of points which are correctly assigned to clusters. If the size of selected training data is too small or too big, the performance of further classification will be affected. Besides this, the quality of selected training data determines the effectiveness of the further classification. In this section, we introduce distance-based, entropy-based training data selection approaches, and propose a self-adaptive labeling method involving distance and entropy to select training data from the clustering results.

\subsubsection{Distance-based Training Data Labeling Algorithm}
The distance-based training data selection is depicted in Algorithm 1.

\begin{algorithm}
\caption{Distance-based Training Data Labeling}
\KwIn{A source data set $\{x_{1},x_{2},...,x_{N}\}$}
\KwOut{Selected training data $E$}
\begin{enumerate}
  \item Initiate $E$ to be the empty set for selected training data\;
  \item Apply clustering algorithm on the given data to get clusters $\{C_1, C_2,...,C_N\}$\;
  \item Calculate the distance of $x_{i}$ to the center of the cluster it belongs to\;
  \item Select the points which are close to the cluster centers with percentage $A\%$ into $E$.
\end{enumerate}
\label{algorithm1}
\end{algorithm}

\subsubsection{Entropy-based Training Data Labeling Algorithm}
First let us introduce the definition of entropy.

\begin{definition}
The information entropy $H(x_{i})$ of point $x_{i}$ is defined as:
\begin{equation}
H(x_i )=-\Sigma_{l=1}^K{P(C_l |x_i)}log_2{P(C_l |x_i)}
\label{eq:eq1}
\end{equation}
where $\Sigma_{l=1}^K P(C_l |x_i)=1$, $P(C_l |x_i)\geq 0, l=1,...,K$ and $P(C_l |x_i)log_2{P(C_l |x_i)}=0$ when $P(C_l ©¦x_i )=0$.
\end{definition}
The information entropy specified in Definition 1 makes clustering get a better data partition. The underlying idea of entropy-based training data selection algorithm is below: if the probability of assigning a point to a cluster is much higher than to other clusters, then its information entropy to the cluster is much lower than to any other clusters. If the probability of assigning a point to several clusters are similar, then it indicates that the point is likely on the boundaries of the clusters.

On top of the above idea, the entropy-based training data selection is given in Algorithm 2.
\begin{algorithm}
\caption{Entropy-based Training Data Labeling}
\KwIn{A source data set $\{x_{1},x_{2},...,x_{N}\}$}
\KwOut{Selected training data $E$}
\begin{enumerate}
    \item Initiate $E$ to be the empty set for selected training data\;
    \item Apply clustering algorithm on the given data to get clusters $\{C_1, C_2,...,C_N\}$\;
    \item Calculate the information entropy of point $x_i$ by (\ref{eq:eq1}) \;
    \item Select points which have the lowest information entropy with percentage $A\%$ into $E$.
\end{enumerate}
\label{algorithm2}
\end{algorithm}

\subsubsection{Self-adaptive Training Data Labeling Algorithm}
The two training data selection algorithms described above (distance-based and entropy-based selection algorithms) have corresponding advantages and disadvantages. The distance-based labeling algorithm is effective for the points which are close to the cluster center. By contrast, the entropy-based selection algorithm has a higher accuracy for the points which are on the boundaries of clusters. We propose a novel and self-adaptive selection algorithm integrating the distance-based and the entropy-based selection algorithms by data characteristics.

\begin{definition}
The Silhouette coefficient $s(x_i)$ of point $x_i$ is defined as:
\begin{equation}
s(x_i )=\frac{b(x_i )-a(x_i)}{max({a(x_i ),b(x_i)})}
\label{eq:eq2}
\end{equation}
where, $a(x_i)$ is the average dissimilarity between point $x_i$ and all other points in the same cluster, and $b(x_i)$ is the average dissimilarity between point $x_i$ and all other points in the next nearest cluster.
\end{definition}

Silhouette coefficient is a good method to evaluate how objects in a cluster are closely related, and how distinct or well-separated a cluster is from other clusters. A higher Silhouette coefficient score relates to a model with better defined clusters.
For a specific cluster, we use the mean silhouette coefficient which is defined in (\ref{eq:eq3}) to evaluate the cluster.

\begin{equation}
MS(C_l)=  \frac{1}{n} \Sigma_{i=1}^n s(x_i)
\label{eq:eq3}
\end{equation}

High $MS(C_{l})$ denotes that the cluster $C_{l}$ is separated from other clusters well and the related objects in this cluster are close with each other. We know that the distance-based training data labeling method is effective for the points which are close to the center of the cluster, the entropy-based training data labeling method outperforms the former on the boundaries of the clusters. The mean silhouette coefficient can be applied to determine which method should be chosen for labeling training data. The integrated self-adaptive training data labeling algorithm is described as Algorithm 3.
\begin{algorithm}
\caption{Self-Adaptive Labeling}
\KwIn{A source data set $\{x_1,x_2,...,x_N\}$}
\KwOut{Selected training data $E$}
\begin{enumerate}
  \item Initiate $E$ to be the empty set for selected training data\;
  \item Apply clustering algorithm on the given data to get clusters $\{C_1, C_2,...,C_N\}$\;
  \item Compute the mean silhouette coefficient $MS(C_l)$ of cluster $C_l$\;
  \item \eIf {$MS(C_l) > threshold$}{Select $A\%$ of points in $C_l$ into $E$ by distance-based labeling algorithm\;}{Select $A\%$ of points in $C_l$ into $E$ by entropy-based labeling algorithm.}
\end{enumerate}

\label{algorithm3}
\end{algorithm}

\section{Integrated Algorithms}

Different supervised classification methods can be applied in the CSAL framework. In this section, we firstly introduce a classification expectation maximization (CEM) algorithm \cite{Cel92-CEM}, then detail our proposed innovative classification algorithm (CSAL) integrating clustering and supervised normal mixture model, followed by three derived algorithms.

\subsection{CEM Algorithm}

The CEM algorithm calculates the parameters, determines the clusters by a classification approach and starts from an initial partition.

\textit{Start:} Given an initial partition $\{x_1,x_2,...,x_N\}$,

\textit{E-step:} For cluster $l=1,...,K$, and data point $i=1,...,N$, calculate the current posterior probabilities of $x_i$ belonging to $C_l$:

\begin{equation}
{\omega_{il}}^k=\frac{\hat{\alpha}_l^k f(x_i,\hat{\mu_l^k},\hat{\Sigma}_t^k)}{\Sigma_{t=1}^K \hat{\alpha}_t^k  f(x_i,\hat{\mu}_l^k,\hat{\Sigma}_t^k)}
\label{eq:eq4}
\end{equation}
where $f(x_i,\hat{\mu_l^k},\hat{\Sigma}_t^k)$ denotes the $d$-dimensional normal density in terms of mean $\hat{\mu_l^k}$ and covariance matrix $\hat{\Sigma}_t^k$.

\textit{C-step:} Assign $x_{i}$ to cluster $C_{l}$ with the maximum posterior probability $\omega_{il}^k$, and generate the partition results $P_k$.

\textit{M-step:} For $l=1,...,K$, calculate the estimates of maximum likelihood $\hat{\alpha}_l^{k+1}$,$\hat{\mu}_l^{k+1}$,$\hat{\Sigma}_t^{k+1}$ by (\ref{eq:eq5}).

\begin{equation}
\left\{\begin{array}{l} \hat{\alpha}_l^{k+1}=\frac{\#P_l^k}{N} \\
\hat{\mu}_l^{k+1}=\frac{\Sigma_{x_i\in P_l^k}x_i}{\#P_l^k} \\
\hat{\Sigma}_t^{k+1}=\frac{\Sigma_{i=1}^K\Sigma_{x_i\in P_l^k}\|x_i-\hat{\mu}_l^{k+1}\|}{Nd}
\end{array} \right.
\label{eq:eq5}
\end{equation}
where $\#P_l^k$ is the number of points assigned to cluster $C_l$.

\subsection{CSAL Algorithm}

The CSAL algorithm is described as follows. First, a clustering algorithm is applied to cluster the data set, and each point is given a class label. Second, $A\%$ data points in each cluster are selected as training data, then a supervised normal mixture model for classification is trained on the selected training data. $\hat{\theta_l^0} = \frac{1}{N}\Sigma_{i=1}^N y_{il}$ are calculated by (\ref{eq:eq6}):

\begin{equation}
\left\{\begin{array}{l} \hat{\alpha}_l^0=\frac{1}{N} \Sigma_{i=1}^N y_{il} \\
\hat{\mu}_l^0=\frac{\Sigma_{i=1}^N y_{il}x_i}{\Sigma_{i=1}^N y_{il}} \\
\hat{\Sigma}_l^0=\frac{\Sigma_{i=1}^N y_{il}(x_i-\hat{\mu}_l^0)(x_i-\hat{\mu}_l^0)^T}{\Sigma_{i=1}^N y_{il}}
\end{array} \right.
\label{eq:eq6}
\end{equation}
Finally, we repeat the process of training data labeling and classification of the whole data set until the convergence of the algorithm. Below, we show the main process of the CSAL algorithm:

\textit{Start:} Given an initial partition $\{x_1,x_2,...,x_N\}$,

\textit{E-step:} For $l=1,...,K$, and $i=1,...,N$, compute the posterior probabilities of $x_i$ belonging to $C_l$:

\begin{equation}
{\hat{\omega}_{il}}^k=\frac{\hat{\alpha}_l^k |\hat{\Sigma}_l^k|^{-\frac{1}{2}} exp{-\frac{1}{2}(x_i-\hat{\mu}_l^k) \hat{\Sigma}_l^k (x_i-\hat{\mu}_l^k)^T}}{\Sigma_{t=1}^K \hat{\alpha}_t^k |\hat{\Sigma}_l^k|^{-\frac{1}{2}} exp{-\frac{1}{2}(x_i-\hat{\mu}_t^k) \hat{\Sigma}_t^k (x_i-\hat{\mu}_t^k)^T}}
\label{eq:eq7}
\end{equation}

\textit{C-step:} Assign $x_i$ to cluster $C_l$ with the maximum posterior probability $\omega_{il}^k$, and set $y_{il}^k=1$ and $y_{it}^k=0,t \neq l$.

\textit{S-step:} Select $A\%$ data points from each cluster by training data selection algorithm. If point $x_i$ is selected as training data, $s_i = 1$; otherwise $s_i = 0$. Let

\begin{equation}
\lambda= \left\{ \begin{array}{ll} 1, & y_{il} = 1, and  \quad s_i = 1;\\
0, & otherwise.\\
\end{array}\right.
\end{equation}

\textit{M-step:} $l=1,...,K$, calculate the estimates of maximum likelihood $\hat{\alpha}_l^{k+1}$, $\hat{\mu}_l^{k+1}$, $\hat{\Sigma}_l^{k+1}$ by (\ref{eq:eq9}).

\begin{equation}
\left\{ {\begin{array}{l}
{\hat \alpha _l^{k + 1} = \frac{{\mathop \sum \nolimits_{i = 1}^N {\rm{\lambda }}_{il}^k}}{{\mathop \sum \nolimits_{t = 1}^K \mathop \sum \nolimits_{i = 1}^N {\rm{\lambda }}_{il}^k}}}\\
{\hat \mu _l^{k + 1} = \frac{{\mathop \sum \nolimits_{i = 1}^N {\rm{\lambda }}_{il}^k{x_i}}}{{\mathop \sum \nolimits_{i = 1}^N {\rm{\lambda }}_{il}^k}}}\\
{\hat \Sigma _l^{k + 1} = \frac{{\mathop \sum \nolimits_{i = 1}^N {y_{il}}({x_i} - \hat \mu _l^{k + 1}){{({x_i} - \hat \mu _l^{k + 1})}^T}}}{{\mathop \sum \nolimits_{i = 1}^N {\rm{\lambda }}_{il}^k}}}
\end{array}} \right.
\label{eq:eq9}
\end{equation}

The CSAL algorithm extends the CEM algorithm: the S-step is added after the C-step to select training data, and in the M-step the parameters are calculated on the selected training data. Different clustering algorithms such as K-means, FCM and Gaussian Mixture Model (GMM)\cite{Carl-GMM} can be integrated with the CSAL algorithm, and respectively generate the derived algorithms as Kmeans-CSAL, FCM-CSAL and GMM-CSAL.

\subsection{CSAL Convergence and Complexity Analysis}

The convergence of the CEM algorithm has been proved \cite{Cel92-CEM}. Similarly, we prove the convergence of the CSAL algorithm as follows.

\textbf{Theorem 1} The log likelihood of CSAL is $\log p\left( {x{|}{{\hat \lambda }^k},{{\hat \alpha }^k},{{\hat \mu }^k},{{\hat \Sigma }^k}} \right)$, and the estimates of the parameters ${\hat \theta ^{k + 1}} = \{ \hat \alpha _l^{k + 1},\hat \mu _l^{k + 1},\hat \Sigma _l^{k + 1}\} _{l = 1}^K$ of the CSAL algorithm converges to fixed values.

\textit{Proof}  The log likelihood is

\begin{equation}
\log \left( {{\rm{p}}\left( {{\rm{x|\theta }}} \right)} \right) = \mathop \sum \limits_{i = 1}^N \mathop \sum \limits_{l = 1}^K {\lambda _{il}}\log {\alpha _l}p({x_i}|{\mu _l},{\Sigma _l})
\label{eq:eq10}
\end{equation}
${\hat \theta ^{k + 1}} = \{ \hat \alpha _l^{k + 1},\hat \mu _l^{k + 1},\hat \Sigma _l^{k + 1}\} _{l = 1}^K$ is the maximum log likelihood of \\ $\log \left( {{\rm{p}}\left( {{\rm{x|}}{{\hat \lambda }^k},{{\hat \alpha }^k},{{\hat \mu }^k},{{\hat \Sigma }^k}} \right)} \right)$, we have:

\begin{equation}
\log \left( {{\rm{p}}\left( {{\rm{x|}}{{\hat \lambda }^k},{{\hat \alpha }^{k + 1}},{{\hat \mu }^{k + 1}},{{\hat \Sigma }^{k + 1}}} \right)} \right) \ge \log \left( {{\rm{p}}\left( {{\rm{x|}}{{\hat \lambda }^k},{{\hat \alpha }^{k}},{{\hat \mu }^{k}},{{\hat \Sigma }^{k}}} \right)} \right)
\label{eq:eq11}
\end{equation}
and when $y_{il}^k = 1$, $\hat w_{il}^{k + 1} \ge \hat w_{it}^{k + 1}$ holds for all $t \ne l$, which implies
\begin{equation}
{\hat \alpha _l}^{k + 1}p({x_i}|{\hat \mu ^{k + 1}},{\Sigma ^{k + 1}}) \ge {\hat \alpha _l}^kp({x_i}|{\hat \mu ^k},{\Sigma ^k})
\label{eg:eg12}
\end{equation}
In addition, the proportion of selected data is the same in each iteration, and the selected data has the highest $\hat w_{il}^{k + 1}$, so

\begin{equation}
\log {\rm{p}}\left( {{\rm{x|}}{{\hat \lambda }^{k + 1}},{{\hat \alpha }^{k + 1}},{{\hat \mu }^{k + 1}},{{\hat \Sigma }^{k + 1}}} \right) \ge \log p\left( {x{\rm{|}}{{\hat \lambda }^k},{{\hat \alpha }^k},{{\hat \mu }^k},{{\hat \Sigma }^k}} \right)
\label{eg:eg13}
\end{equation}
Since the number of instances into each cluster is finite, $\log p\left( {x{\rm{|}}{{\hat \lambda }^k},{{\hat \alpha }^k},{{\hat \mu }^k},{{\hat \Sigma }^k}} \right)$ converges to a fixed value. If $k$ is large enough, then
\begin{equation}
\left\{ {\begin{array}{*{20}{c}}
{{{\hat \alpha }^k} = {{\hat \alpha }^{k + 1}}}\\
{{{\hat \mu }^k} = {{\hat \mu }^{k + 1}}}\\
{{{\hat \Sigma }^k} = {{\hat \Sigma }^{k + 1}}}
\end{array}} \right.
\label{eg:eg14}
\end{equation}
In each iteration, the complexity of the CSAL algorithm is $O(dKN)$.

\section{Experiments and Evaluation}

In this section, we firstly introduce the experiment settings and evaluation method. Then we detail the experimental results.

\subsection{Data Sets}

The data sets mainly involve two synthetic data sets and four real data sets from the UC Irvine Machine Learning Repository \cite{UCI-Frank+Asuncion}.

\textit{1) Synthetic Gaussian data:} The Gaussian data sets are synthesized in two ways. Firstly, 200 instances (we name the data set GData1) falling into two classes of bivariate Gaussian density with the following parameter settings:

\begin{equation}
\left\{ {\begin{array}{*{20}{c}}
{{\mu _1} = \left( {1,1} \right)}\\
{{\mu _2} = \left( {2,0} \right)}\\
{{\Sigma _1} = \left[ {\begin{array}{*{20}{c}}
1\\
0
\end{array}\begin{array}{*{20}{c}}
0\\
{0.25}
\end{array}} \right]}\\
{{\Sigma _2} = \left[ {\begin{array}{*{20}{c}}
{0.8}\\
0
\end{array}\begin{array}{*{20}{c}}
0\\
1
\end{array}} \right]}
\end{array}} \right.
\label{eg:eg15}
\end{equation}
Secondly, 300 instances (we call it GData2) falling into three classes of bivariate Gaussian density with the following parameter settings:

\begin{equation}
\left\{ {\begin{array}{*{20}{c}}
{{\mu _1} = \left( {0,0} \right)}\\
{{\mu _2} = \left( {6,6} \right)}\\
{{\mu _3} = \left( {-10,-10} \right)}\\
{{\Sigma _1} = \left[ {\begin{array}{*{20}{c}}
1\\
0
\end{array}\begin{array}{*{20}{c}}
0\\
{1}
\end{array}} \right]}\\
{{\Sigma _2} = \left[ {\begin{array}{*{20}{c}}
{3}\\
0
\end{array}\begin{array}{*{20}{c}}
0\\
3
\end{array}} \right]}\\
{{\Sigma _3} = \left[ {\begin{array}{*{20}{c}}
100\\
0
\end{array}\begin{array}{*{20}{c}}
0\\
{100}
\end{array}} \right]}\\
\end{array}} \right.
\label{eg:eg16}
\end{equation}

\textit{2) Real Data Sets:} Four real data sets from UCI repository, which are Iris, Heart Diseases, New Thyroid and Wine are exploited in this paper.

\subsection{Experimental Settings}

Classification accuracy is used to evaluate the performance of the CSAL algorithms.
\begin{equation}
ClassificationAccuracy = {{\mathop \sum \nolimits_{l = 1}^K {D_l}} \over N}
\label{eg:eg18}
\end{equation}
where $D_l$ is the number of samples correctly partitioned in the genuine class $C_l$.

To evaluate the performance of our proposed CSAL algorithms, experiments are conducted as follows.
First, we compare different training data labeling strategies which connect clustering algorithms with supervised classifiers together. Second, we compare the derived CSAL algorithms with three different clustering algorithms (K-means, FCM and GMM). Next, we evaluate the derived CSAL algorithms comparing with clustering algorithms integrating different classification methods (Naive bayes and SVM). Then they are compared with the CEM algorithms. Last, the execution time of all the algorithms are compared.

\subsection{Experimental Results}

\subsubsection{Training Data Labeling}
\begin{figure}[htp]
\centerline{\includegraphics[scale=0.4]{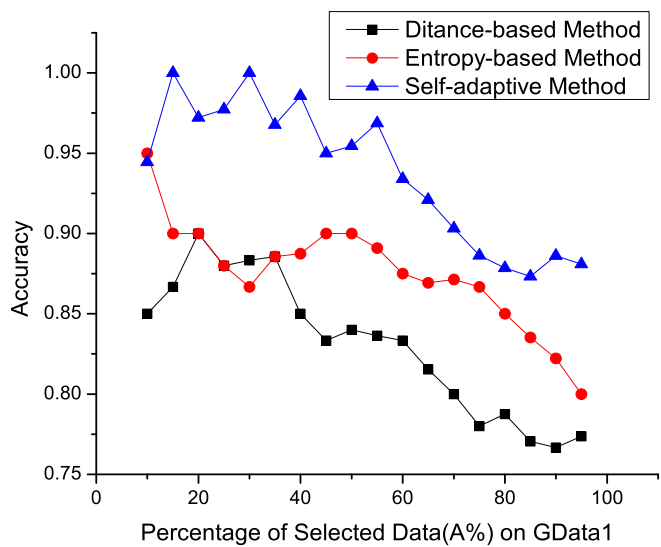}} \caption{The Classification Accuracy of Self-adaptive, Entropy-based, Distance-based Methods for Training Data Labeling on GData1} \label{fig_gdata1}
\end{figure}

\begin{figure}[htp]
\centerline{\includegraphics[scale=0.4]{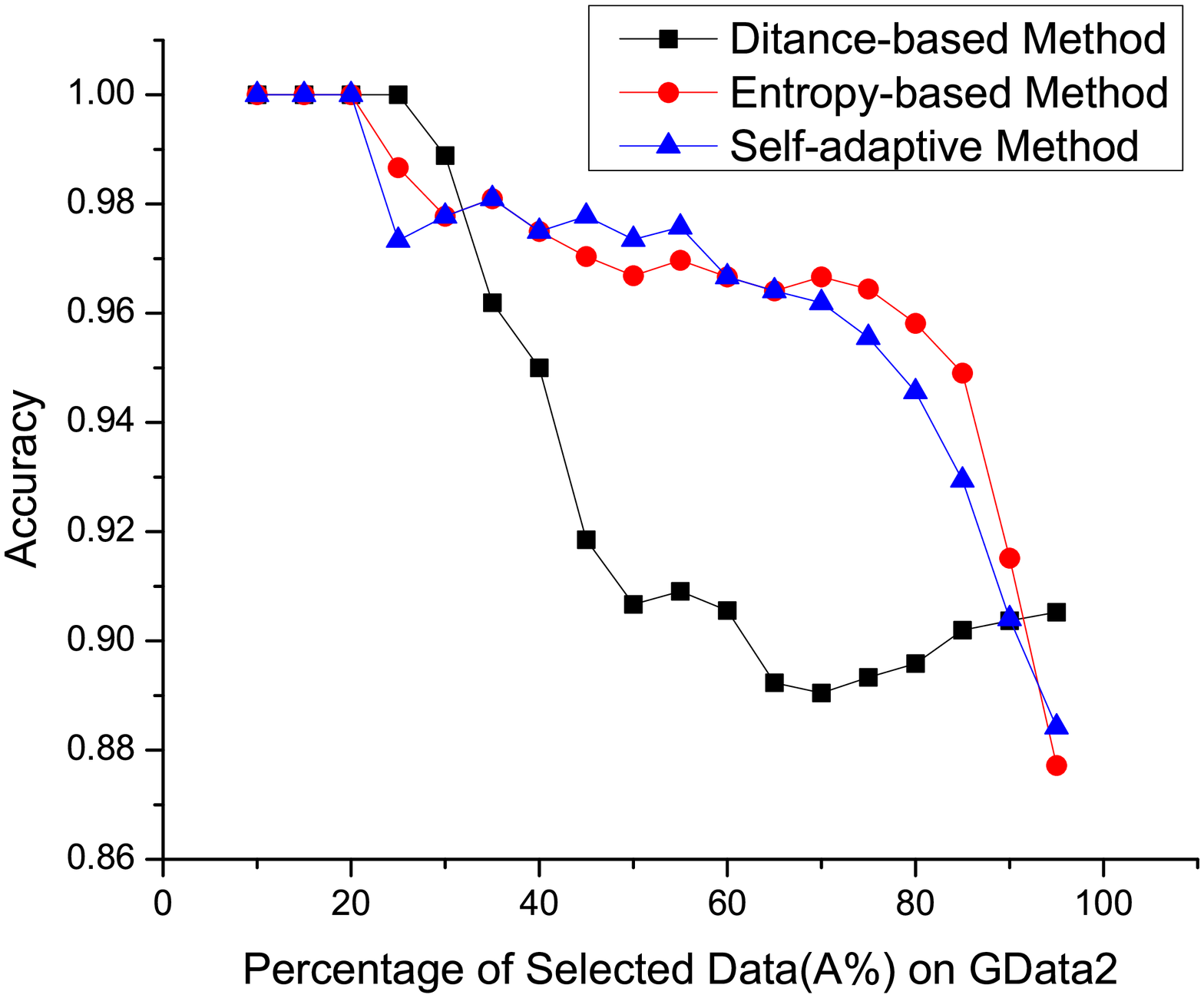}} \caption{The Classification Accuracy of Self-adaptive, Entropy-based, Distance-based Methods for Training Data Labeling on GData2} \label{fig_gdata2}
\end{figure}

\begin{figure}[htp]
\centerline{\includegraphics[scale=0.4]{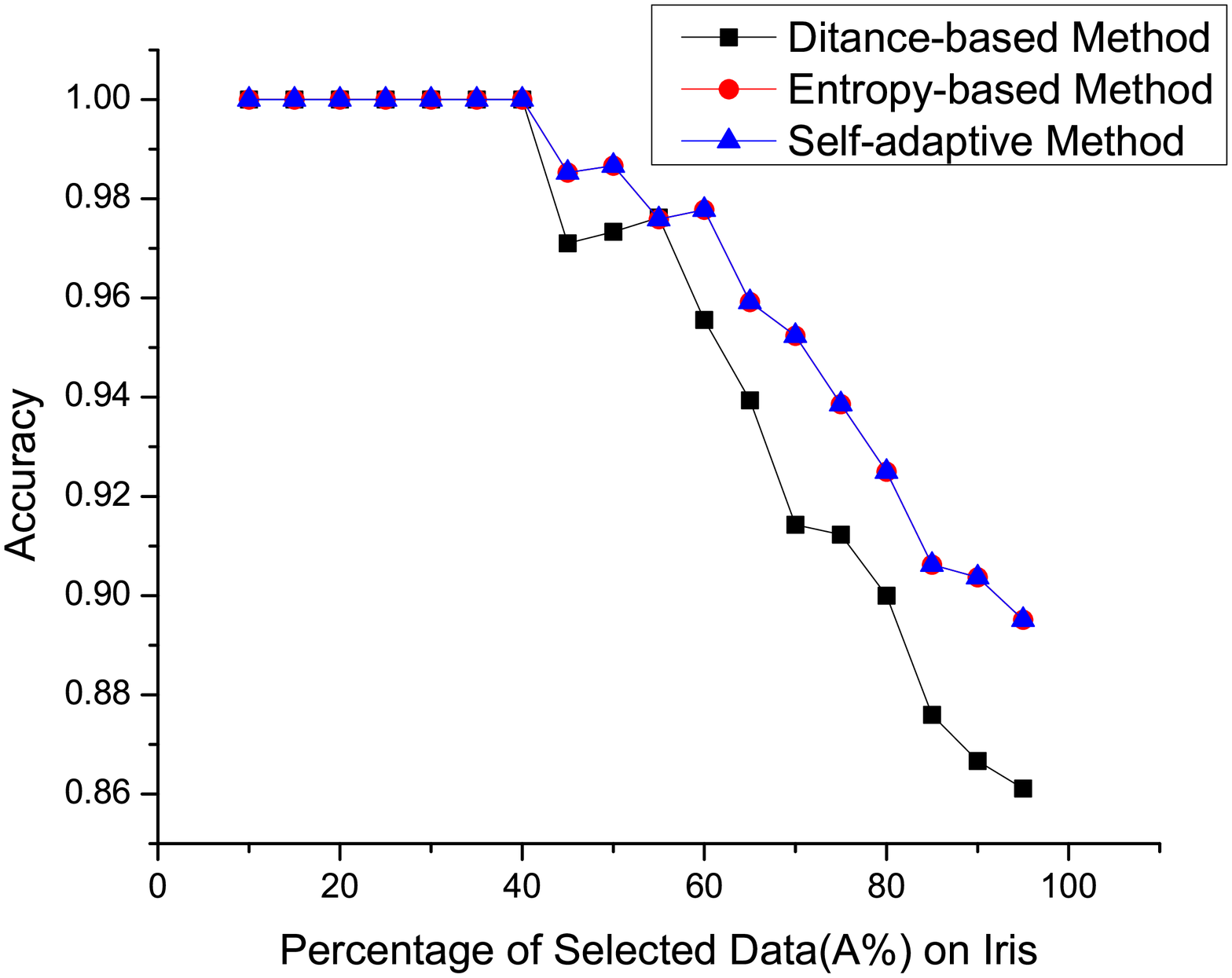}} \caption{The Classification Accuracy of Self-adaptive, Entropy-based, Distance-based Methods for Training Data Labeling on Iris} \label{fig_iris}
\end{figure}

\begin{figure}[htp]
\centerline{\includegraphics[scale=0.4]{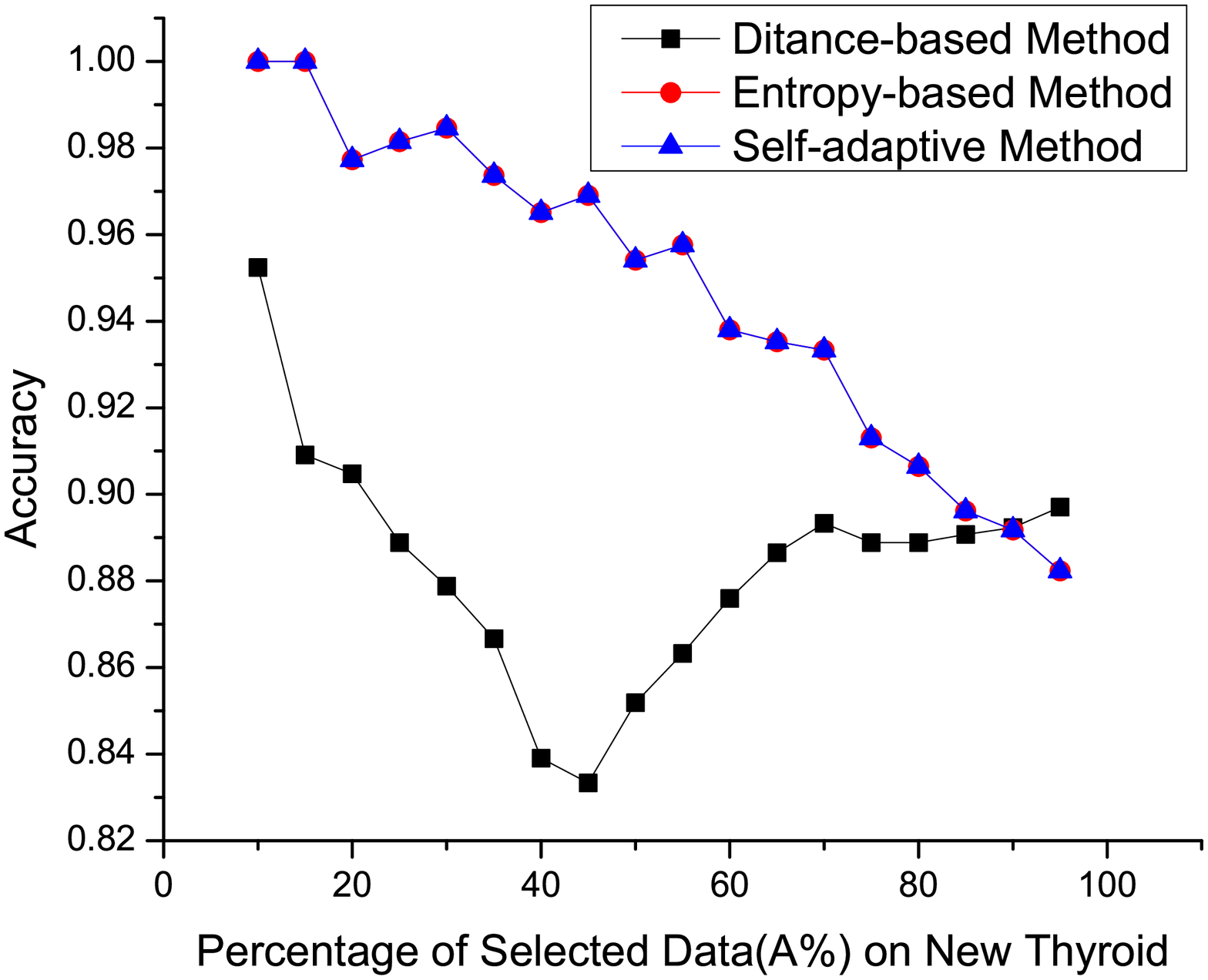}} \caption{The Classification Accuracy of Self-adaptive, Entropy-based, Distance-based Methods for Training Data Labeling on New Thyroid} \label{fig_newThroid}
\end{figure}

\begin{figure}[htp]
\centerline{\includegraphics[scale=0.4]{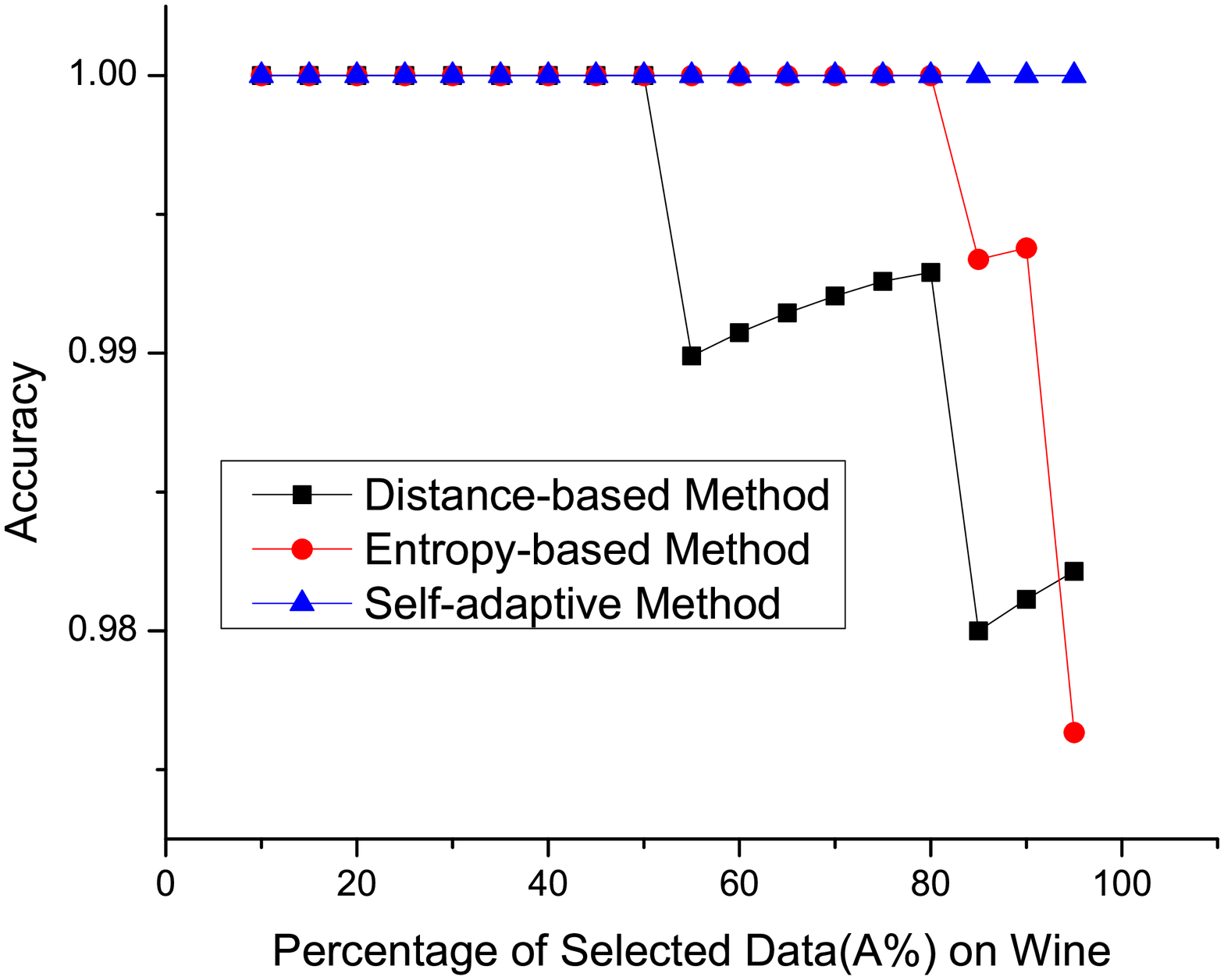}} \caption{The Classification Accuracy of Self-adaptive, Entropy-based, Distance-based Methods for Training Data Labeling on Wine} \label{fig_wine}
\end{figure}

\begin{figure}[htp]
\centerline{\includegraphics[scale=0.4]{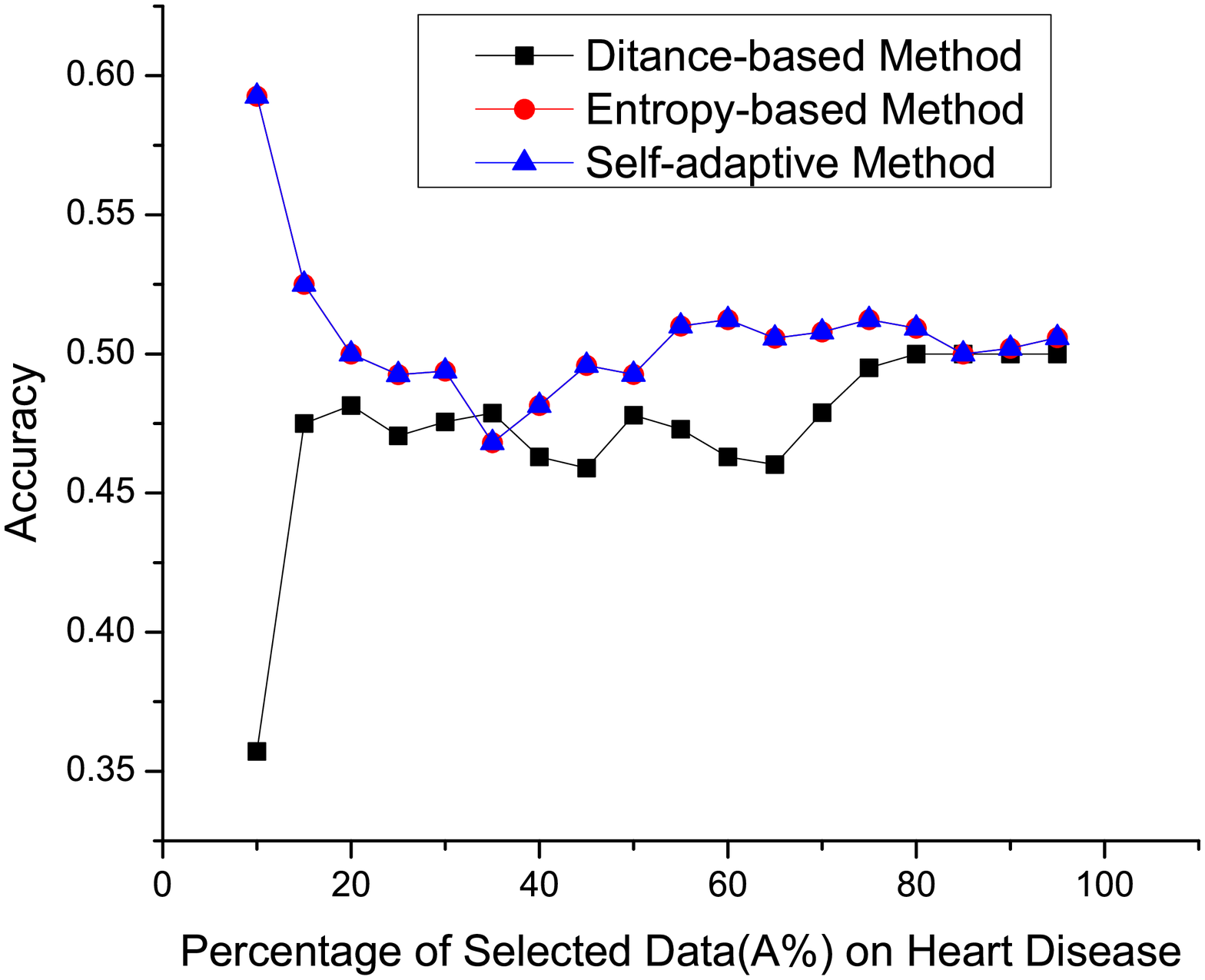}} \caption{The Classification Accuracy of Self-adaptive, Entropy-based, Distance-based Methods for Training Data Labeling on Heart Disease} \label{fig_heart}
\end{figure}

In this paper, different training data labeling strategies (distance-based, entropy-based and self-adaptive methods) are compared, where $threshold$ is empirically set to 0.35. Fig. \ref{fig_gdata1}-\ref{fig_heart} show that the entropy-based method performs better than the distance-based method for selecting training data, and self-adaptive method outperforms the distance-based and entropy-based methods. Because entropy-based and self-adaptive methods have the same performance on Iris and New Thyroid data sets, their curves are overlapped in Figs. \ref{fig_iris} and \ref{fig_newThroid}.

\subsubsection{Comparison of the CSAL with Traditional Clustering Algorithms}
\begin{figure}
\centerline{\includegraphics[scale=0.4]{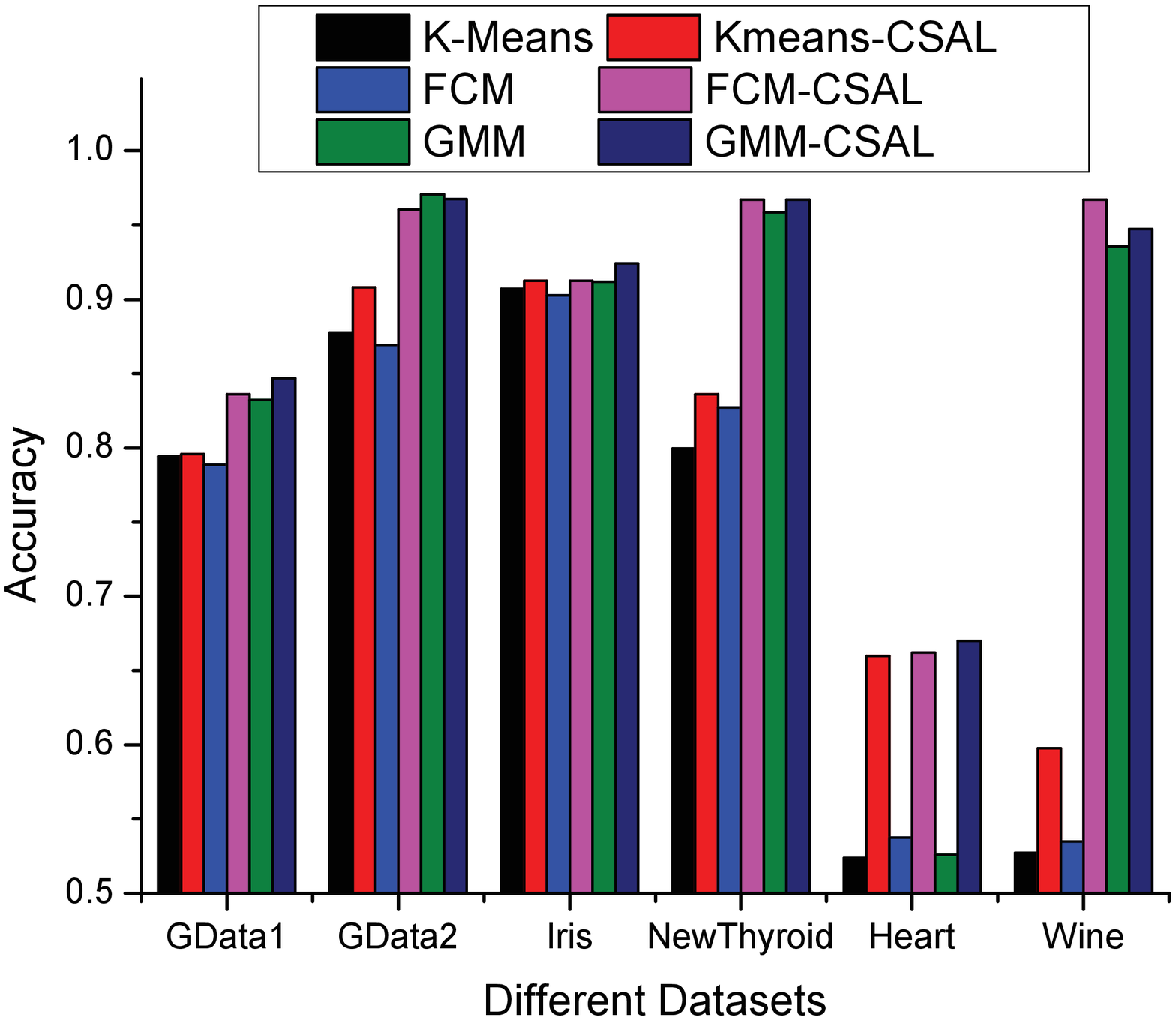}} \caption{Performance Comparison of K-means vs Kmeans-CSAL, FCM vs FCM-CSAL, GMM vs GMM-CSAL} \label{compare1}
\end{figure}

In order to evaluate our CSAL framework, we compare it with traditional clustering algorithms (K-means, FCM and GMM). The comparison result is described in Fig. \ref{compare1} which clearly indicates that the CSAL is more effective than traditional clustering algorithms. The derived Kmeans-CSAL, FCM-CSAL and GMM-CSAL algorithms performs much better than the corresponding clustering algorithms.

\subsubsection{Comparison of the CSAL Algorithms with Traditional Clustering algorithms Integrating Different Supervised Classification Methods}
\begin{figure}[htp]
\centerline{\includegraphics[scale=0.4]{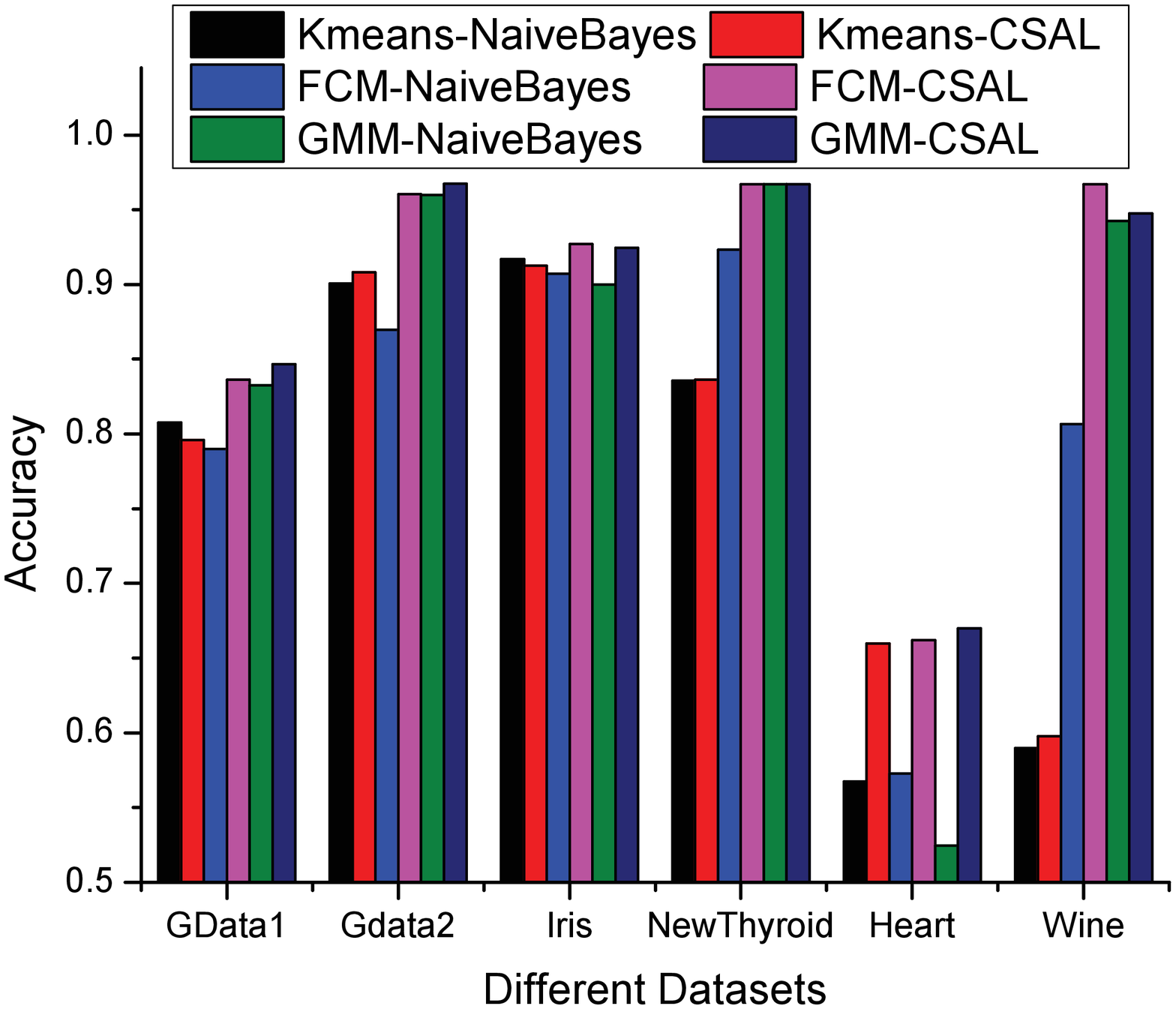}} \caption{Performance Comparison of Kmeans-NaiveBayes vs Kmeans-CSAL, FCM-NaiveBayes vs FCM-CSAL, GMM-NaiveBayes vs GMM-CSAL} \label{compare2}
\end{figure}

\begin{figure}[htp]
\centerline{\includegraphics[scale=0.4]{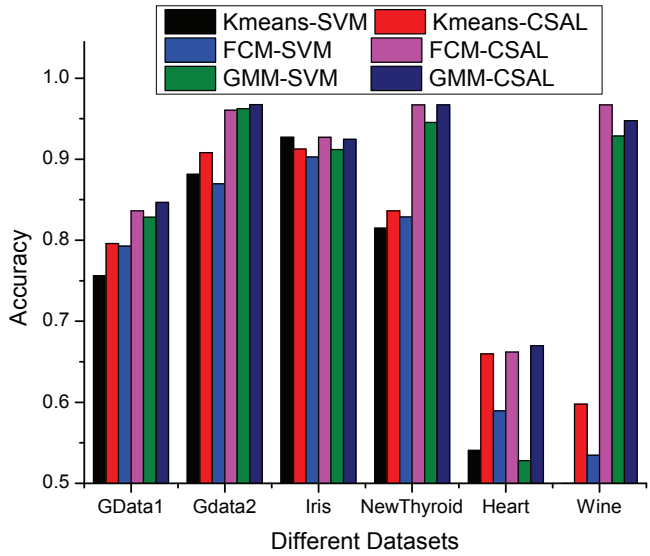}} \caption{Performance Comparison of Kmeans-SVM vs Kmeans-CSAL, FCM-SVM vs FCM-CSAL, GMM-SVM vs GMM-CSAL} \label{compare3}
\end{figure}

The comparison results are shown in Figs. \ref{compare2} and \ref{compare3}. For all data sets, Fig. \ref{compare2} indicates that the derived CSAL algorithms are more effective than the corresponding clustering algorithms which integrate Naive Bayes classifier without the training data labeling process. For most of the data sets, Fig. \ref{compare3} shows that the derived CSAL algorithms perform better than the corresponding clustering algorithms which integrate SVM classifier without a training data labeling process.

\subsubsection{Comparison with the CEM algorithm}
\begin{figure}[htp]
\centerline{\includegraphics[scale=0.4]{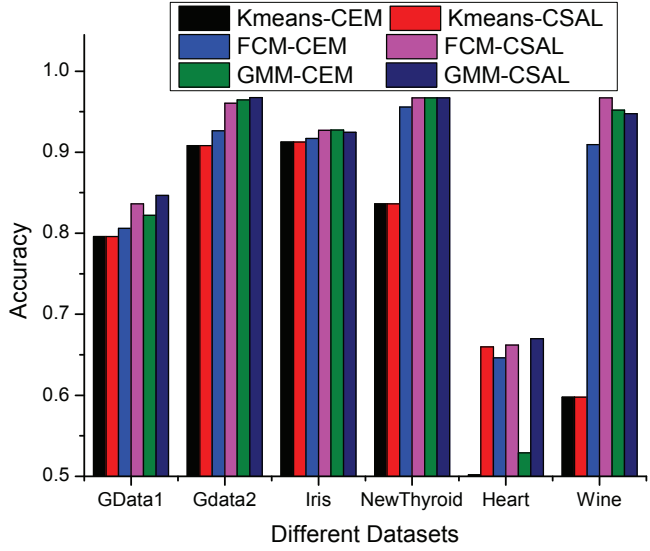}} \caption{Performance Comparison of Kmeans-CEM vs Kmeans-CSAL, FCM-CEM vs FCM-CSAL, GMM-CEM vs GMM-CSAL} \label{compare4}
\end{figure}

The CSAL is also compared with the CEM algorithm, which is shown in Fig. \ref{compare4}. It illustrates that the derived CSAL algorithms are more effective than the corresponding CEM algorithms.

\subsubsection{Comparison of Runtime}
The following Table \ref{table2} indicates the comparison of execution time of different algorithms. The algorithms are coded in Matlab and executed in a machine with 3.10GHz CPU and 4.00GB RAM. Table 2 shows that the execution time of the CSAL process is longer than that of the Naive Bayes classifier and the CEM algorithms, but much shorter than that of the SVM algorithm. The runtime of GMM is much longer than the CSAL process, consequently GMM-CSAL takes slightly longer time than GMM. The runtime of the Kmeans-CSAL algorithm is in the same size of that of K-means. The runtime of the FCM-CSAL algorithm is much longer than that of FCM, but they are still in the same order of magnitude.

\begin{table*}[htbp]\caption{The Execution Time (in seconds)}
\centering
\begin{tabular}{|c|c|c|c|c|c|c|}
\hline
Algorithms & GData1 & GData2 & Iris & Heart & NewThyroid & Wine\\ \hline
K-means  & 0.1391 & 0.1644 & 0.1405 & 0.2772 & 0.2777 & 0.1645 \\ \hline
Kmeans-NaiveBayes & 0.1475 & 0.1748 & 0.1482 & 0.2873 & 0.2876 & 0.1746 \\ \hline
Kmeans-SVM & 0.2095 & 0.3166 & 0.2395 & 0.8544 & 0.4467 & 0.3848 \\ \hline
Kmeans-CEM & 0.1494 & 0.1802 & 0.1495 & 0.2892 & 0.2951 & 0.1827 \\\hline
Kmeans-CSAL & 0.1839 & 0.2212 & 0.1672 & 0.3315 & 0.3335 & 0.2212 \\ \hline
GMM & 2.3863 & 4.7656 & 2.5647 & 1.56 & 1.1084 & 1.7350 \\ \hline
GMM-NaiveBayes & 2.3925 & 4.7746 & 2.5709 & 1.5694 & 1.1196 & 1.7467 \\ \hline
GMM-SVM & 2.4752 & 4.9741 & 2.6884 & 2.2133 & 1.3220 & 2.0992  \\ \hline
GMM-CEM & 2.3958 & 4.7823 & 2.5794 & 1.5735 & 1.1276 & 1.7548 \\ \hline
GMM-CSAL & 2.4184 & 4.8218 & 2.6034 & 1.6052 & 1.1782 & 1.7903 \\ \hline
FCM	 & 0.0088 &	0.0158 &	0.0123 &	0.0173 &	0.06 &	0.0423  \\ \hline
FCM-NaiveBayes &	0.0181 &	0.0275 &	0.0227 &	0.0314 &	0.0714 &	0.0547 \\  \hline
FCM-SVM &	0.165 &	0.4844 &	0.1998 &	1.1267 &	0.7006 &	0.4798 \\ \hline
FCM-CEM &	0.0214 &	0.0344 &	0.0292 &	0.0339 &	0.0789 &	0.0617 \\ \hline
FCM-CSAL &	0.0681 &	0.0881 &	0.0631 &	0.0955 &	0.1288 &	0.1125 \\ \hline
\end{tabular}
\label{table2}
\end{table*}

\section{Conclusion}
The challenging problem in clustering and classification is the absence of labeled data, however, it is costly and time-consuming to obtain labeled data in real applications. In this paper, we proposed a novel clustering framework CSAL to solve the challenge. In this framework, we also proposed a new self-adaptive labeling approach for training data selection and compared it with distance-based and entropy-based labeling methods. The experiments on publicly data sets showed that the self-adaptive labeling method outperforms the distance-based and entropy-based methods. Additionally, the experiments also demonstrated that the CSAL is more effective than the corresponding comparison partners. From the experiments we know that the CSAL can handle unlabeled data well, but it still does not consider the data characteristics such as gaussian or gamma distribution of the data. We believe that the performance of the CSAL framework can be further enhanced if we involve the data distribution in the future.

\bibliography{ref}
\bibliographystyle{plain}

%\begin{acknowledgements}
%If you'd like to thank anyone, place your comments here
%and remove the percent signs.
%\end{acknowledgements}

% BibTeX users please use one of
%\bibliographystyle{spbasic}      % basic style, author-year citations
%\bibliographystyle{spmpsci}      % mathematics and physical sciences
%\bibliographystyle{spphys}       % APS-like style for physics
%\bibliography{}   % name your BibTeX data base

\end{document}